\documentclass{article}
\usepackage{spconf,amsmath,graphicx}
\usepackage{mathrsfs}
\usepackage{float}
\usepackage{amssymb}
\usepackage{cite}
\usepackage{booktabs}
\usepackage{multicol,listings,multirow}
\usepackage{lipsum}
\usepackage{tikz}
\usepackage{amssymb}
\usepackage{wrapfig}
\usepackage{array}
\usepackage{makecell}
\usepackage{booktabs}
\usepackage{array}
\usepackage{color}
\usepackage{makecell}
\usepackage[colorlinks=true,
            linkcolor=blue,
            filecolor=blue,      
            urlcolor=blue,       
            citecolor=blue
            ]{hyperref}
\usepackage[marginal]{footmisc}
\usepackage[misc]{ifsym}
\usepackage{soul}
\usepackage{enumitem}
\setlist{nosep, leftmargin=14pt}

\usepackage{mwe} 

\title{Text-guided Foundation Model Adaptation for Long-Tailed Medical Image Classification}
\name{Sirui Li$^{1\ddagger}$, Li Lin$^{1,2\ddagger}$, Yijin Huang$^{1,3}$, Pujin Cheng$^{1,2}$, Xiaoying Tang$^{1,4*}$
\thanks{$^\ddagger$ {These authors contributed equally}.}}
\address{$^1$Department of Electronic and Electrical Engineering, Southern University of Science and Technology,\\ Shenzhen, China\\
$^2$Department of Electrical and Electronic Engineering, The University of Hong Kong,\\Hong Kong SAR, China\\
$^3$School of Biomedical Engineering, University of British Columbia,\\ Vancouver, Canada \\
$^4$Jiaxing Research Institute, Southern University of Science and Technology,\\ Jiaxing, China}
\begin{document}
%
\maketitle

\begin{abstract}
In medical contexts, the imbalanced data distribution in long-tailed datasets, due to scarce labels for rare diseases, greatly impairs the diagnostic accuracy of deep learning models. Recent multimodal text-image supervised foundation models offer new solutions to data scarcity through effective representation learning. However, their limited medical-specific pretraining hinders their performance in medical image classification relative to natural images. To address this issue, we propose a novel \textbf{T}ext-guided \textbf{F}oundation model \textbf{A}daptation for \textbf{L}ong-\textbf{T}ailed medical image classification \textbf{\textit{(TFA-LT)}}. We adopt a two-stage training strategy, integrating representations from the foundation model using just two linear adapters and a single ensembler for balanced outcomes. Experimental results on two long-tailed medical image datasets validate the simplicity, lightweight and efficiency of our approach: requiring only 6.1\% GPU memory usage of the current best-performing algorithm, our method achieves an accuracy improvement of up to 27.1\%, highlighting the substantial potential of foundation model adaptation in this area.
\end{abstract}
\vspace{-0.2cm}
\section{Introduction}
\vspace{-0.2cm}
\label{sec:intro}

\begin{figure}[htbp]
\centering
\vspace{-0.2cm}
\setlength{\abovecaptionskip}{-0.2cm}   
\setlength{\belowcaptionskip}{-0.7cm}   
\centerline{\includegraphics[height=4.7cm]{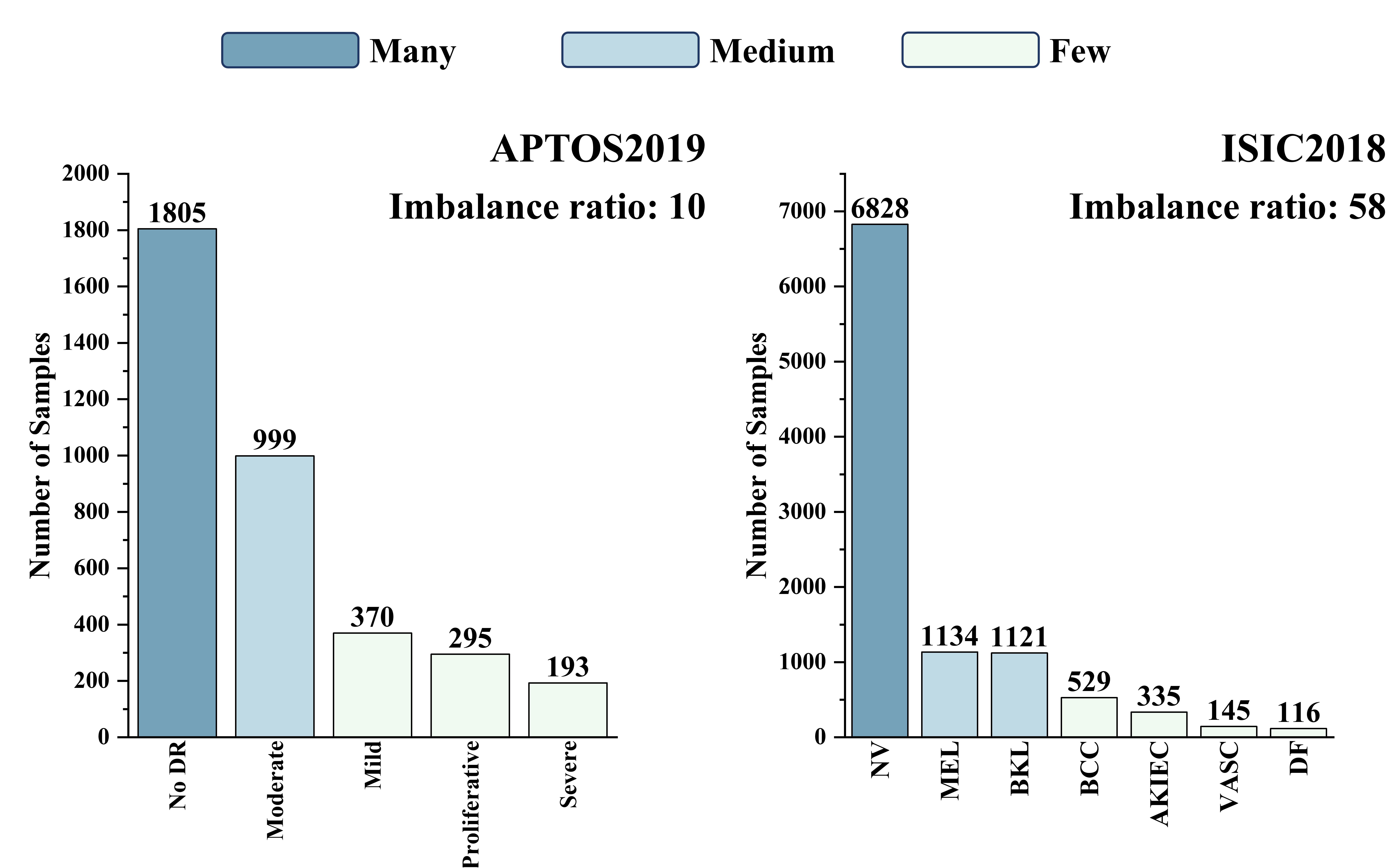}}
\vspace{-0.3cm}
\caption{Class distributions and subset divisions of two long-tailed medical datasets employed in our experiments.}
\medskip
\vspace{-0.7cm}
\label{fig1}
\end{figure}

Benefiting from large-scale, balanced, and high-quality labeled data, deep neural networks have achieved remarkable accomplishments in the field of medical image classification~\cite{cai2022uni4eye,proco, hyjssit}. However, real-world medical datasets, often skewed due to challenges in annotating rare diseases, result in models biased towards common classes, impacting diagnosis of crucial rare conditions.

To alleviate this issue, prior approaches mainly focus on class re-balancing, information augmentation and module improvement. Class re-balancing strategies, such as re-sampling~\cite{resampling, resample2} and re-weighting~\cite{reweighting, reweighting2}, aim for balanced class distribution. However, most methods boost tail class performance at the cost of head class performance. Information augmentation techniques like data augmentation~\cite{ccmix,balancedmixup} and knowledge distillation~\cite{dkd,medkd} infuse extra information during training to enhance long-tailed efficacy. Despite the improvement, extra time and prior knowledge might be required. Furthermore, aiming at improving the network structure for long-tailed learning, many works concentrate on classifier design~\cite{classifier} and feature extractor~\cite{extractor} enhancement. Even though impressive performance gains, high computational needs limit their downstream application.

Recently, foundation models like CLIP~\cite{clip} have made significant strides in computer vision. These models, pre-trained on vast image-text pairs and using contrastive loss to align image and text representations, excel in image classification, particularly in zero-shot contexts. In few-shot scenarios, a linear classifier trained on image encoder outputs often outperforms many algorithms. However, their effectiveness in medical image classification is limited as most do not include medical-related image-text pairs in pretraining. Meanwhile, our experiments indicate that a linear classifier trained solely on the representations output from image encoder does not achieve satisfactory performance on long-tailed datasets. While there are some foundation models~\cite{medclip,flair} pre-trained on medical images currently available, they tend to focus on single modalities, compromising their multimodal efficacy. Furthermore, compiling a comprehensive set of medical image-text pairs for a universal multimodal medical foundation model is cost-prohibitive.

Given these challenges, we aim to propose a novel framework, namely \textbf{T}ext-guided \textbf{F}oundation model \textbf{A}daptation for \textbf{L}ong-\textbf{T}ailed medical image classification \textbf{\textit{(TFA-LT)}}. By leveraging the richer relational representations in the text space, as opposed to one-hot labels, we enhance the foundation model's performance in medical long-tailed classification. Our training process comprises two stages. In stage I, with the foundation model's encoders fixed, we train two linear adapters on resampled training sets. Using an adjustable hyperparameter $\lambda$ and dynamic residual connections, we align visual representations with textual representations to achieve better knowledge embedding. Stage II involves freezing the initial models and combining them at either logit or feature level with a linear ensembler, creating more comprehensive and balanced representations. We employ focal loss to aid the ensembler in fine-tuning on the un-resampled training set.

The main contributions of this paper are three-fold: (1) We design an adapter to align and balance visual and textual representations, thereby achieving more optimized knowledge embedding. (2) We introduce a novel two-stage training strategy, requiring only two linear adapters and one linear ensembler to integrate balanced representations at the logit or feature level. (3) We demonstrate the simplicity and efficiency of our approach through extensive tests on two medical datasets, improving overall accuracy by up to 27.1\% over the current best method while using only 6.1\% of its GPU memory, revealing the immense potential of foundation model adaptation in long-tailed tasks. The source code is available at: \url{https://github.com/sirileeee/TFA-LT}.

\vspace{-0.2cm}
\section{Method}
\label{sec:method}
\begin{figure*}[htbp]
\centering
\vspace{-1.4cm}
\setlength{\abovecaptionskip}{-0.2cm}   
\setlength{\belowcaptionskip}{-0.8cm}   
\centerline{\includegraphics[height=9.5cm]{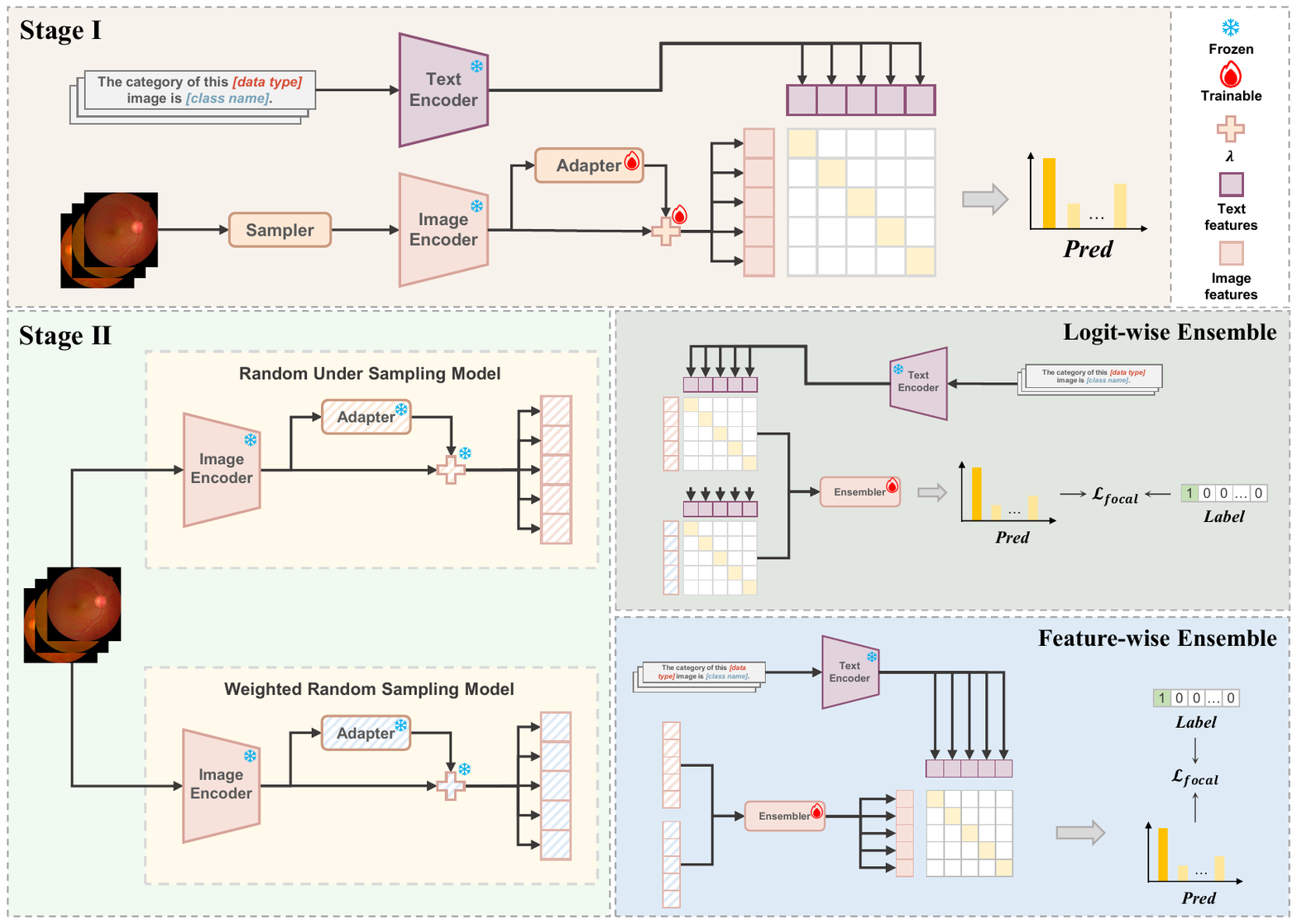}}
\vspace{-0.3cm}
\caption{The architecture of our two-stage framework: the upper section depicts stage I and the design of our residual connection adapter, while the lower section outlines stage II and the two levels of representation ensemble.}
\vspace{-0.5cm}
\label{fig2}
\end{figure*}

\vspace{-0.1cm}
The overall framework, including the adapter design and the two-stage training process, is shown in Fig.~\ref{fig2}.

\vspace{-0.2cm}
\subsection{Residual Connection Adapter}
\label{subsec:adapter}

We utilize a linear adapter and a residual connection hyperparameter $\lambda$ to adjust the relative representations in the visual space, enabling better alignment with textual representations. The initial visual feature $f_v'$ with dimension $d_v$ is produced from the image $I$ to be classified through the freeze pre-trained image encoder $E_v$: $f_v'=E_v(I)\in\mathbb{R}^{d_v}$. Inspired by~\cite{ballad}, we keep the visual encoder freeze and train a linear adapter $A\in\mathbb{R}^{d_v \times d_v}$ to generate enhanced and balanced visual representation $h_v$: $h_v=A(f_v')\in\mathbb{R}^{d_v}$. To strengthen stability, we employ a learnable hyperparameter $\lambda$  to create a residual connection between the original feature $f_v'$ and the adapter feature $h_v$. To regulate the bounds of $\lambda$, we subject it to a sigmoid transformation $\sigma$ to it throughout the computational process. The final visual feature $f_v$ is defined as:
\vspace{-0.2cm}
\begin{equation}
f_v = \sigma(\lambda) f_v' + [1 - \sigma(\lambda)] h_v,
\end{equation}

\vspace{-0.5cm}
\subsection{Representation Ensemble Training}
\label{subsec:2stage}

\subsubsection{Balanced Visual Representation}
\label{subsubsec:1st-stage}
To help generate a balanced visual representation, we first train two residual connection adapters on two re-balanced training sets in parallel. We generate these two datasets through Weighted Random Sampling (WRS) and Random Under Sampling (RUS). WRS assigns higher sampling probabilities to less-represented classes, promoting the selection of rare samples, while RUS randomly eliminates instances from the head classes. Both methods promote balanced training and representation acquisition.

For annotated text $T$, we replace one-hot labels with semantically richer prompts, using `The category of this \textit{[data type]} image is \textit{[class name]}.' to better align representations across different sample modalities. The text feature $f_t$ with dimension $\mathbb{R}^{d_t}$ obtained from the frozen pre-trained text encoder $E_t$ could be represented as: $f_t=E_t(T)\in\mathbb{R}^{d_t}$.

Foundation models like CLIP are equipped with visual and text encoders to generate highly aligned embeddings in a unified feature space, guiding the image classification process by computing the similarity between the feature vectors of input images and various text prompts. For a long-tailed dataset with $C$ classes, adapter $A$ and hyperparameter $\lambda$ optimize the visual representation by maximizing the cosine similarity between visual feature $f_v$ and text feature $f_t$:
\vspace{-0.2cm}
\begin{equation}
p(\hat{y}=c\mid I) = \frac{\text{exp}(\text{sim}(f_t^c, f_v)/\tau)}{\sum_{c'=1}^C \text{exp}(\text{sim}(f_t^{c'}, f_v)/\tau)},
\end{equation}
\vspace{-0.3cm}

where $\tau$ is a temperature parameter and $\text{sim}(\cdot,\cdot)$ refers to cosine similarity.

\vspace{-0.3cm}
\subsubsection{Representation Integration}
\label{subsubsec:2nd-stage}

After completing the training in the stage I, we freeze the adapter and $\lambda$ based on our initial training settings. We then process the original, long-tailed training set through both models to derive visual features. Considering the overfitting issues in head classes and underfitting in tail classes that resampling techniques might introduce, in stage II we utilize a learnable linear ensembler to integrate the visual representations from the two models, aiming to produce a balanced and information-rich representation for long-tailed classification. We propose two representation integration methods: logit-wise ensemble and feature-wise ensemble.

We propose that ensembling at the feature level allows the model to integrate all relevant information in a more profound and nuanced manner. Meanwhile, given that logits typically have lower dimension and more consistent distribution, the logit level ensemble is closer to the model's output and may therefore more directly influence the model's final decision. We simply need to adjust the dimension of the ensembler to match the dimension of the logits or features, allowing for easy integration of visual representations. We use $f_v^w$ and $l_v^w$ to represent the visual features and logits from the WRS training set, and $f_v^r$, $l_v^r$ for those from the RUS training set. The linear ensembler is denoted as $K$. Overall, the output logits $z$ could be presented as:
\vspace{-0.1cm}
\begin{equation}
z = 
\left\{
\begin{array}{lc}
\text{sim}(K(f_v^w, f_v^r), f_t)/\tau~, & feature-wise \\
K(l_v^w, l_v^r)~, & logit-wise
\end{array}
\right..
\end{equation}
\vspace{-0.3cm}

To promote fair class representation learning by the ensembler, we utilize focal loss~\cite{focal} for fine-tuning.

\vspace{-0.1cm}
\section{Experiments}
\vspace{-0.1cm}
\label{sec:exp}

\begin{table*}[htbp]
\renewcommand\arraystretch{1}
\centering
\resizebox{0.9\linewidth}{!}
{
\small
\setlength{\abovecaptionskip}{-0.2cm}   
\begin{tabular}{c|ccccc|ccccc}
\specialrule{0.10em}{0pt}{0pt}
\multicolumn{1}{c|}{\multirow{2}{*}{\textbf{Algorithm}}} & \multicolumn{5}{c|}{\textbf{APTOS2019}}         & \multicolumn{5}{c}{\textbf{ISIC2018}}         \\
\multicolumn{1}{l|}{}                           & \textbf{Many} & \textbf{Med} & \textbf{Few} & \textbf{Total} & \textbf{F1} & \textbf{Many} & \textbf{Med} & \textbf{Few} & \textbf{Total} & \textbf{F1}    \\
\specialrule{0.05em}{0pt}{1pt}
\specialrule{0.05em}{1pt}{0pt}
CE (baseline)                                              & 95.33             & 61.33 & 35.11 & 52.39 & 50.05   & \textbf{88.33}              & 34.17             & 40.00             & 45.24             & 44.05                       \\
Focal loss~\cite{focal}                                     & 95.33             & 58.67 & 38.67 & 52.80 & 50.59     & \underline{86.67}              & 57.50             & 38.73             & 50.95             & 47.51                     \\
ResLT~\cite{reslt}                                         & \textbf{96.03}             & \textbf{88.09} & 13.29 & 44.80 & 38.72     & \underline{86.67}     & 36.67             & 45.83             & 49.05             & 41.95                     \\
PaCo~\cite{paco}                                          & 77.37             & 12.60 & 58.23 & 52.93 & 46.62        & 5.00              & 25.83             & 57.93 & 41.19             & 34.47                  \\
GPaco~\cite{gpaco}                                         & 89.33             & 12.87 & 59.93 & 56.40 & 49.36   & 23.33              & 39.17             & \textbf{86.67}    & 64.05             & 62.27                       \\
BALLAD~\cite{ballad}                                      & \textbf{96.03}             & 32.67 & 47.11 & 54.00 & 51.87     & 66.83              & 61.94             & 62.16             & 62.86             & 62.22                      \\
BCL~\cite{bcl}                                             & \underline{96.00}             & 52.67 & 49.56 & 59.47 & 59.99   & 85.00              & \underline{63.33}    & 71.67             & \textbf{71.19}    & \textbf{71.01}             \\
\specialrule{0.05em}{0pt}{1pt}
\specialrule{0.05em}{1pt}{0pt}
CLIP~\cite{clip} zero-shot                                 & 0.00              & 2.00  & 20.00 & 20.40 & 7.49    & 5.00               & 37.50             & 20.00             & 22.86             & 17.89                       \\
CLIP~\cite{clip} linear-prob                               & \underline{96.00}             & \underline{85.33} & 16.67 & 46.27 & 40.83  & \textbf{88.83}  & 51.67             & 28.33             & 44.05             & 42.98                        \\
\textbf{TFA-LT (logit)}                                     & \underline{96.00}             & 58.67 & \underline{69.78} & \underline{72.80} & \underline{72.36}  & 65.01              & \textbf{66.67} & 73.33             & 70.24             & \underline{70.47} \\
\textbf{TFA-LT (feature)}                                   & 90.00             & 68.67 & \textbf{73.11} & \textbf{75.60} & \textbf{75.63}  & 73.23              & \underline{63.33}             & \underline{73.41}             & \underline{70.48} & 70.45               \\
\specialrule{0.10em}{0pt}{0pt}
\end{tabular}
}
\caption{Comparisons with state-of-the-art methods. The best ones are \textbf{bolded} while the second best ones are \underline{underlined}.}
\label{tab:main}
\vspace{-0.3cm}
\end{table*}

\subsection{Datasets and Evaluation}
\label{subsec:dataset}
We conduct our experiments on two public long-tailed medical datasets: ISIC2018~\cite{isic2018}, a large-scale collection of dermoscopy images, and APTOS2019~\cite{aptos2019}, a fundus dataset from the APTOS2019 blindness detection competition. To ensure a fair and unbiased performance evaluation across all classes, unaffected by the imbalanced distribution, we split both dataset into an 8:2 training-test ratio, maintaining a balanced distribution of classes within the test set. We further divide each dataset into three subsets: many, med(ium), and few, based on the number of samples per class, to thoroughly assess the effectiveness of our method across head and tail classes. During evaluation, we calculate the overall accuracy and f1-score of the dataset, as well as the accuracy for each subset. Dataset distribution and division are detailed in Fig.~\ref{fig1}.

\vspace{-0,2cm}
\subsection{Implementation}
\label{subsec:imp}
We conduct experiments on NVIDIA TITAN RTX GPUs using Pytorch, utilizing the pre-trained CLIP ViT/B-16 as both image and text encoders, which remain unchanged during training. We apply SGD with a momentum of 0.9 and weight decay of $5e^{-4}$ as our optimizer, and set the batch size to 128. The initial learning rate is set to 0.1 with cosine annealing to ensure the model's convergence. We let our approach train for 200 epochs in stage I and 100 epochs in stage II, while allowing all other comparison methods to train for 300 epochs. To mitigate randomness, all experiment results are averaged on 3 random seeds.

\vspace{-0.2cm}
\subsection{Results}
\label{subsec:result}

We evaluate TFA-LT in comparison to 7 widely-used and state-of-the-art long-tailed classification algorithms and also the two classification approaches, zero-shot and linear-prob, used by CLIP. As tabulated in Table~\ref{tab:main}, the results indicate that with just 200 epochs of training in stage I and 100 epochs in stage II, both our logit-wise and feature-wise ensemble strategies notably surpass other state-of-the-art methods on APTOS2019 and achieve comparable results to top performers on ISIC2018.

\begin{figure}[htbp]
\centering
\vspace{-0.2cm}
\setlength{\abovecaptionskip}{-0.2cm}   
\setlength{\belowcaptionskip}{-0.5cm}   
\centerline{\includegraphics[height=5.0cm]{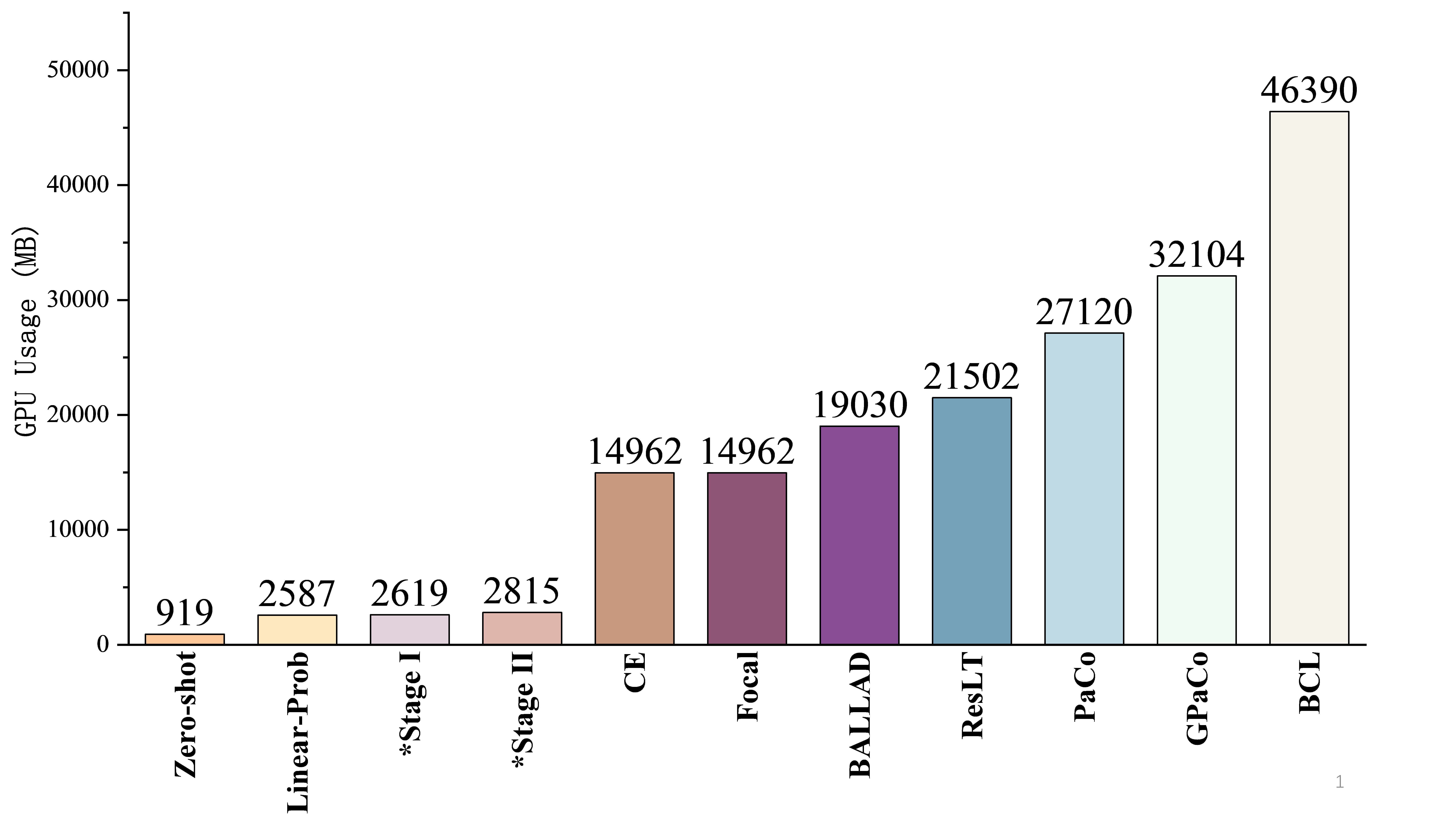}}
\vspace{-0.4cm}
\caption{Comparison of GPU usage during training between TFA-LT's two stages, prefixed with *, and the other 9 comparative benchmark methods during training.}
\medskip
\vspace{-0.3cm}
\label{fig3}
\end{figure}

In order to more explicitly demonstrate the lightweight and usability of our method, we list the GPU memory usage of our approach compared to the aforementioned 9 frameworks during training in Fig.~\ref{fig3}. We note that our method demands solely 2.56GB of GPU memory in stage I and 2.75GB in stage II, significantly less than the resource usage of all other comparative methods. Although our approach only yield suboptimal results on the ISIC2018 dataset, it requires a mere 6.1\% of the computational resources needed for the best-performing method BCL. Conversely, on the APTOS2019 dataset, our method outperforms the current best-performing technique, achieving a 27.1\% increase in total accuracy, establishing a leading position.

\vspace{-0.5cm}
\subsection{Ablation Studies}
\label{subsec:abla}
We further investigate the effectiveness of the residual connection adapter in stage I and the representation ensembler in stage II. CLIP zero-shot is selected as our baseline, since its image and text encoders are fixed, eliminating the need for additional component training. We first integrate a residual connection adapter into the visual backbone without balancing the training set. Subsequently, we employ logit-wise and feature-wise ensemblers for integrating the visual representations. Finally, we demonstrate that the two-stage training strategy we propose, which combines the residual connection adapter and the representation ensembler, can maximize the potential of both modules, resulting in significant performance improvements. The results can be found in Table~\ref{tab:ablation}.

\begin{table}[htbp]
\renewcommand\arraystretch{1.0}
\centering
\resizebox{0.85\linewidth}{!}
{
\small
\begin{tabular}{cc|cc|cc}
\specialrule{0.10em}{0pt}{0.5pt}
\multirow{2}{*}{\textbf{Adapter}} & \multirow{2}{*}{\textbf{Ensembler}} & \multicolumn{2}{c|}{\textbf{APTOS2019}} & \multicolumn{2}{c}{\textbf{ISIC2018}} \\
                                  &                                    & \textbf{Total}      & \textbf{F1}      & \textbf{Total}      & \textbf{F1}      \\
\specialrule{0.05em}{0pt}{0pt}
-                                 & -                                  & 20.40               & 7.49             & 22.86               & 17.89            \\
\checkmark                        & -                                  & 61.60               & 58.88            & 66.19               & 63.36            \\
-                                 & logit-wise                              & 39.21               & 31.01            & 34.52               & 32.42            \\
-                                 & feature-wise                            & 51.47               & 42.67            & 67.86               & 66.96            \\
\checkmark                        & logit-wise                              & \underline{72.80}               & \underline{72.36}            & \underline{70.24}               & \textbf{70.47}   \\
\checkmark                        & feature-wise                            & \textbf{75.60}      & \textbf{75.63}   & \textbf{70.48}      & \underline{70.45}            \\
\specialrule{0.10em}{0pt}{0pt}
\end{tabular}
}
\vspace{-0.1cm}
\caption{Ablation studies of our framework. The best outcomes are highlighted in \textbf{bold}, while the second best ones are marked with \underline{underline}.}
\label{tab:ablation}
\end{table}

\vspace{-0.5cm}
\section{Conclusion}
\label{sec:typestyle}
\vspace{-0.1cm}
In this paper, we purpose \textbf{T}ext-guided \textbf{F}oundation model \textbf{A}daptation for \textbf{L}ong-\textbf{T}ailed medical image classification \textbf{\textit{(TFA-LT)}}, which leverages the power of pre-trained foundation model to address the problem of long-tailed medical image classification. Adopting a two-stage training strategy for balanced representation ensemble, TFA-LT demands only 6.1\% of the GPU memory required by BCL, the leading algorithm in long-tailed classification. Through training two linear adapters and one linear ensembler, it achieves state-of-the-art level performance on two long-tailed medical datasets. The simplicity, lightweight, and efficiency of TFA-LT shed light on a new potential approach for leveraging foundation models to address long-tailed challenges.

\vspace{-0.2cm}
\section{Acknowledgments}
\label{sec:acknowledgments}
\vspace{-0.2cm}
This study was supported by the Shenzhen Basic Research Program (JCYJ20190809120205578); the National Natural Science Foundation of China (62071210); the Shenzhen Science and Technology Program (RCYX20210609103056042); the Shenzhen Basic Research Program (JCYJ2020092515384\\7004); the Shenzhen Science and Technology Innovation Committee (KCXFZ2020122117340001).

\small
\bibliographystyle{IEEEbib}
\bibliography{refs}

\end{document}